# A Smart Healthcare System for Monkeypox Skin Lesion Detection and Tracking

Huda Alghoraibi[1+], Nuha Alqurashi[1+], Sarah Alotaibi[1+], Renad Alkhudaydi[1+], Bdoor Aldajani[1+], Lubna Alqurashi[1+], Jood Batweel[1+], Maha A. Thafar[1*]

[1] College of Computers and Information Technology, Computer Science Department,
Taif University, Taif, Saudi Arabia.

[+] All authors contributed equally.

[*] Corresponding author: (e-mail: m.thafar@tu.edu.sa).

**Abstract**

Monkeypox is a viral disease characterized by distinctive skin lesions and has been reported in many countries. The recent global outbreak has emphasized the urgent need for scalable, accessible, and accurate diagnostic solutions to support public health responses.

In this study, we developed ITMA'INN, an intelligent, AI-driven healthcare system specifically designed to detect Monkeypox from skin lesion images using advanced deep learning techniques. Our system consists of three main components. First, we trained and evaluated several pretrained models using transfer learning on publicly available skin lesion datasets to identify the most effective models. For binary classification (Monkeypox vs. non-Monkeypox), the Vision Transformer, MobileViT, Transformer-in-Transformer, and VGG16 achieved the highest performance, each with an accuracy and F1-score of 97.8%. For multiclass classification, which contains images of patients with Monkeypox and five other classes (chickenpox, measles, hand-foot-mouth disease, cowpox, and healthy), ResNetViT and ViT Hybrid models achieved 92% accuracy, with F1 scores of 92.24% and 92.19%, respectively. The best-performing and most lightweight model, MobileViT, was deployed within the mobile application. The second component is a cross-platform smartphone application that enables users to detect Monkeypox through image analysis, track symptoms, and receive recommendations for nearby healthcare centers based on their location. The third component is a real-time monitoring dashboard designed for health authorities to support them in tracking cases, analyzing symptom trends, guiding public health interventions, and taking proactive measures.

This system is fundamental in developing responsive healthcare infrastructure within smart cities. Our solution, ITMA'INN, is part of revolutionizing public health management.

**Keywords**— AI, Monkeypox Detection, Deep learning, CNN, Vision Transformer, Smart Healthcare.

# 1. Introduction

On March 11, 2020, the World Health Organization (WHO) declared the novel coronavirus (COVID-19) outbreak a global pandemic. Calling it "the first pandemic of the 21st century" because it spread quickly from continent to continent, causing more than 8,000 infections in 8 months — with a 10% case fatality ratio (Sandoiu, 2020). Recently, a new public health risk emerged when a multi-country Monkeypox outbreak was reported to the WHO by several non-endemic countries. On September 14, 2022, about 103 member states from six regions reported 59,147 confirmed cases of Monkeypox and 22 deaths. Given the current outbreak of Monkeypox cases, the WHO has declared it a global health emergency ("Multi-country monkeypox outbreak in non-endemic countries: Update," n.d.). Dealing with epidemic diseases such as Monkeypox and limiting their spread is critical to maintaining the public health of the entire population. Relying only on traditional clinical examinations can be time-consuming, not easily accessible, and cause an overload on the medical professionals and hospitals. Therefore, there is a pressing need for an intelligent, effective, and scalable diagnostic system for Monkeypox disease to support early detection, reduce transmission, and assist both individuals and healthcare providers (Sorayaie Azar et al., 2023).

Recent advances in artificial intelligence (AI), particularly in machine learning (ML), deep learning (DL), and computer vision, have enabled promising solutions across various domains, especially in medicine and healthcare (Alamro et al., 2023; Albaradei et al., 2023, 2021; Nasr et al., 2021; Thafar et al., 2023; Zhang et al., 2025; Zhang and Boulos, 2023). These technologies have played a significant role in automating medical image analysis tasks, such as object detection, image segmentation, and image classification (Li et al., 2024; Tsuneki, 2022). In particular, convolutional neural networks (CNNs) and transformer-based models have demonstrated impressive performance in classifying various skin lesions, including skin cancer detection and dermatological disease recognition (De et al., 2024; Yolcu Oztel, 2024). However, despite the growing urgency for effective diagnostic tools, the application of these models to Monkeypox detection remains limited in scope, performance, and real-world deployment.

This research addresses these challenges and proposes an end-to-end smart healthcare system for Monkeypox detection and tracking by integrating DL and other AI and computer vision techniques. The system consists of three key components: First, a DL-based classification model trained on skin lesion images to accurately detect Monkeypox disease. Second, a smart mobile application that deploys an AI-based model for real-time diagnosis and symptom tracking with more extra features. Third, a monitoring dashboard to visually represent Monkeypox cases and their geographical spread detected by the mobile application, allowing health authorities to respond rapidly and prevent disease outbreaks. This system also aims to reduce hospital overload by providing a preliminary, fast, and accurate diagnosis at home through the image-based diagnosis system, thus reducing hospital visits and saving resources for more critical cases. It also limits disease transmission and treats patients as quickly as possible by directing them to the nearest health center if the infection is confirmed, which preserves citizens' public health and safety.

This study is particularly relevant in the context of Saudi Arabia's Vision 2030, which emphasizes the integration of AI and digital health technologies to enhance the efficiency of healthcare services. By aligning with these national objectives, the proposed system aims to support the development of smart health infrastructure within the broader framework of smart cities. Our goal is to offer an accessible, low-cost, and AI-driven solution that enables early diagnosis and supports disease surveillance.

The key contributions of this research are summarized as follows:
- We developed *ITMA'INN*, an AI-powered healthcare system for early detection and monitoring of Monkeypox disease from skin lesion images with efficient, accurate, and fast diagnostic capabilities, supporting the healthcare system and preventing disease transmission.
- The system supports both binary classification (Monkeypox vs. non-Monkeypox) and multiclass classification (Mpox, Chickenpox, Measles, Cowpox, hand-foot-mouth disease (HFMD) and Normal), leveraging fine-tuned pretrained deep learning models including ViT, TNT, Swin Transformer, MobileViT, ViT Hybrid, ResNetViT, VGG16, ResNet50, and EfficientNet-B0.
- We developed and implemented a user-friendly and cross-platform mobile application that enables users to upload skin lesion images for real-time AI-based diagnosis, track symptoms, receive guidance on the nearest healthcare centers, and access relevant and up-to-date disease information.
- We created a real-time monitoring dashboard to support healthcare authorities by visualizing case distribution, tracking patients' trends and patterns, and facilitating data-driven decision-making in response to potential outbreaks.
- This study aligns with Saudi Arabia's Vision 2030 in integrating AI and telemedicine to improve the quality and efficiency of healthcare services in smart cities.

The remainder of this research paper is structured as follows. Section 2 reviews the existing literature on Monkeypox detection using deep learning models. Section 3 describes the datasets utilized and provides a detailed explanation of the proposed methodology. Section 4 outlines the development of the mobile application and the dashboard. Section 5 presents the experimental setup and evaluation metrics. Section 6 discusses the results and key findings. Section 7 concludes the study and highlights potential future directions.

## 2. Related Works

Recently, researchers have paid more attention to developing image-based AI systems to identify Monkeypox and other viral skin-related infections (Debelee, 2023; Singh et al., 2022; Sreekala et al., 2022), especially with the rapid advancement of AI that assists the development of such diagnostic systems (Asif et al., 2024; Chadaga et al., 2023; Patel et al., 2023; Sorayaie Azar et al., 2023). ML, DL, and transfer learning techniques have emerged as promising tools to capture the characteristics of skin lesions, facilitating Monkeypox disease detection (Altun et al., 2023; Jaradat et al., 2023; Nayak et al., 2023; Sorayaie Azar et al., 2023; Thieme et al., 2023).

With the emergence of transfer learning, several studies have applied CNN-based pretrained models for binary and multiclass classification of skin lesion images. One of these studies is the Monkeypox detection using a deep neural network (DNN) (Sorayaie Azar et al., 2023). The authors utilized a dataset consisting of Monkeypox, Chickenpox, Measles, and Normal skin images. They explored two scenarios: a binary classification task (Monkeypox vs. non-Monkeypox) and a multiclass classification task (Monkeypox, Chickenpox, Measles, and Normal skin) to differentiate between skin diseases that exhibit visually similar lesions. They evaluated seven CNN-based pretrained models, which are InceptionResNetV2, InceptionV3, ResNet152V2, VGG16, VGG19, Xception, and DenseNet201. Those models prove their efficiency in image classification tasks. The result showed that the DenseNet201 model outperformed the other models using several evaluation metrics, achieving an accuracy of 97.63% in the two-class scenario and 95.18% in the four-class scenario. They also employed interpretability techniques like Local Interpretable Model-agnostic Explanations (LIME) and Gradient-weighted Class Activation Mapping (Grad-CAM), which helped visualize the image regions that influenced the model's decision. Another recent study published in 2023 (Nayak et al., 2023) employed DL to diagnose Monkeypox from skin lesion images. Using a publicly available dataset, this study evaluated five pretrained CNN models: GoogLeNet, Places365-GoogLeNet, SqueezeNet, AlexNet, and ResNet-18. After performing hyperparameter optimization, ResNet18 obtained the best accuracy of 99.49%. The modified version of the model also performed well, with accuracies above 95%. ResNet18 was particularly effective due to its residual connections, which helped mitigate the vanishing gradient problem and allowed the model to focus on critical image features. LIME was also utilized here to explain the reasoning behind the classifier's predictions. Interestingly, the Vision Transformer (ViT-B16) showed extremely lower performance than CNN-based models on the same dataset. This poor performance highlights the challenges of employing transformer-based models in medical image classification, particularly when the dataset is limited in size or diversity. Nonetheless, in contrast to those findings, our study demonstrates that with suitable preprocessing, data augmentation, and fine-tuning techniques, transformer-based models (i.e., ViT, MobileViT, and TNT) can outperform standard CNNs in both binary and multiclass Monkeypox classification tasks.

Several other studies have extended the investigation of pretrained models through transfer learning, focusing on hyperparameter tuning and real-world deployment, building on earlier CNN-based methods and advanced alternative architectures. For instance, Murat Altun and coauthors (Altun et al., 2023) utilized other advanced and more complex CNN-based pretrained models and applied them in a transfer learning fashion for Monkeypox detection. These models are MobileNetV3-s, EfficientNetV2, ResNET50, VGG19, and DenseNet121. They utilized AUC, accuracy, recall, loss, and F1-score metrics to evaluate and compare the different models. The study investigated several hyperparameter values for optimization, such as the batch size, number of epochs, image size, layer count, activation function, optimizer, and loss function. The optimized hybrid MobileNetV3-s model achieved the best score, with an average F1-score of 0.98, AUC of 0.99, accuracy of 0.96, and recall of 0.97. Despite this study's effectiveness and high performance, it still faced some limitations, including potential overfitting due to limited dataset diversity and the need for further real-world testing to ensure generalizability across different populations and skin types.

Similar to the previous study, Jaradat and coauthors (Jaradat et al., 2023) expanded this approach by incorporating additional CNN-based pretrained models in image-based Monkeypox identification. They investigated five popular CNN pretrained models: MobileNetV2, VGG19, VGG16, ResNet50, and EfficientNetB3, and then fine-tuned them in a transfer learning approach. The results showed that the MobileNetV2 model performed the best, achieving an accuracy of 98.16%, a recall of 0.96, a precision of 0.99, and an F1-score of 0.98. The study highlighted the practicality of these models for rapid and accurate clinical diagnosis, especially when integrated into mobile platforms as a real-time assessment tool to detect Monkeypox cases. Likewise, another study (Thieme et al., 2023) explored the development of DL algorithms to detect Monkeypox from skin images named MPXV-CNN. The MPXV-CNN model was trained using dermoscopic and clinical images, accurately distinguishing between different skin diseases. The MPXV-CNN method employed several CNN architectures, such as ResNet18, ResNet34, ResNet50, ResNet152, DenseNet169, and VGG19_bn, and applied them in a transfer learning style. In the experiment, the authors utilized data augmentation to increase the size and diversity of the training data. The study adopted cross-validation for evaluation. The model was trained on 130,000 images and achieved 90% accuracy. They were able to integrate the MPXV-CNN method into a smartphone application. They developed a free, open-source application called PoxApp that allows users to take photos of lesions, answer questions, and get a risk assessment within five minutes. The app aims to enhance accessibility for communities with limited healthcare resources, encouraging individuals to pursue medical attention. Notably, the app can detect Monkeypox at different stages of the disease and offers users five levels of advice, ranging from "no action required" to "immediate consultation with a doctor". Users can also submit their results to contribute to research efforts, helping scientists predict potential waves in Monkeypox infections and establish an early warning system. The advantages of this technology include increased diagnostic precision, scalability, and rapid screening capabilities. However, it is essential to address potential drawbacks, such as biases stemming from limited or imbalanced datasets, as well as the necessity for further validation across diverse clinical settings.

The last study (Sahin et al., 2022) presented an Android mobile application that utilized DL techniques to analyze skin lesion images. The application has been developed using Android Studio and the Java programming language. The system captures video input, extracts frames, and classifies lesions using several CNN-based pretrained models. Among six evaluated CNN models, the best-performing model was MobileNetv2. Moreover, the application includes the feature of making a quick initial diagnosis and can also classify images with 91.11% accuracy. In addition, the average inference times were observed to range from 19 ms to 831 ms. Despite the success of this application and promising results, the authors discussed some weaknesses, such as the limited size of the Monkeypox skin lesion images in the dataset, which caused an imbalance of dataset issues and thus affected the performance.

Although several methods have been developed for Monkeypox disease detection using different DL techniques, there is still significant room for improvement. Most existing studies have employed CNN-based models, with limited exploration of Vision Transformers and their advanced variants, which are known for their strong performance in image classification tasks. Furthermore, few works have integrated these models into practical, real-time systems that combine accurate detection with mobile accessibility and public health monitoring. This study addresses several gaps by developing an AI-powered solution that leverages state-of-the-art transformer-based pretrained models, a user-friendly mobile application, and a real-time monitoring dashboard to support both individual diagnosis and broader epidemic tracking.

## 3. Methodology

### 3.1 Datasets

In this research, we utilized two publicly available datasets to train and evaluate the DL models for Monkeypox detection and classification. Both datasets were obtained from Kaggle and curated to classify skin lesions associated with Monkeypox infections. To enhance the classification task, improve model generalization, and mitigate overfitting, several classical data augmentation techniques were applied to both datasets. Data augmentation techniques include both geometric and photometric transformations such as rotation, shear, translation, hue/saturation jitter, and elastic deformation to enrich the training data diversity. The augmented data was used only for training, while validation and test sets remained unaltered to ensure fair performance evaluation.

**3.1.1 Binary Classification Dataset**

The binary classification task utilizes the Monkeypox Skin Lesion Dataset (MLSD) (Ali et al., 2022). This dataset consists of two classes: Monkeypox and non-Monkeypox. The dataset is a collection of 228 original images divided into 102 images for the Monkeypox and 126 images for the other classes (i.e., non-Monkeypox). The non-Monkeypox class comprises skin lesion images of chickenpox and measles, selected for their visual similarity to Monkeypox lesions. The dataset is publicly available in the Monkeypox Skin Lesion Dataset and accessible through the Kaggle repository. Both the original and augmented datasets were organized into separate directories to maintain a clear distinction between raw and synthetic data. The statistics of this dataset are shown in Table 1.

**3.1.2 Multiclass Classification Dataset**

For the multiclass classification task, the Mpox Skin Lesion Dataset Version 2.0 (MSLD v2.0) developed by (Ali et al., 2024) were utilized. MSLD v2.0 comprises 755 original skin images collected from 541 distinct patients, providing a diverse and representative sample. MSLD v2.0 dataset is categorized into six classes: Monkeypox (MKP), Chickenpox (CHP), Measles (MSL), Cowpox (CWP), Hand-Foot-Mouth Disease (HFMD), and Healthy skin. According to the original dataset creators, the dataset has received endorsement from professional dermatologists and obtained approval from appropriate regulatory authorities. The dataset is publicly available and accessible through Kaggle Mpox Skin Lesion Dataset Version 2.0. To ensure result reproducibility, the original images and augmented images are organized in different folders. Data augmentation was performed using OpenCV, applying a wide range of transformations, including rotation, translation, reflection, shear, color adjustments, noise addition, sharpening, blurring, elastic deformation, brightness adjustment, and scaling. This approach substantially increased the size and diversity of the datasets and contributed to improved model accuracy. The statistics of the original and augmented data for each class are presented in Table 2.

**Table 1:** Original and augmented images per class in the binary classification dataset (MSLD)

| Image Classes | Original Images | Augmented Images (Training only) |
|---|---|---|
| Monkeypox | 102 | 1148 |
| Non-Monkeypox | 126 | 1414 |
| **Total** | 228 | 2562 |

Table 2: Original and augmented images per class in the multiclass classification dataset (MSLDv2.0)

| Image Classes | Original Images | Augmented Images (Training only) |
|---|---|---|
| Monkeypox | 284 | 1135 |
| Chickenpox | 75 | 1096 |
| Measles | 55 | 1085 |
| Cowpox | 66 | 1119 |
| HFMD | 161 | 1129 |
| Healthy | 114 | 1089 |
| **Total** | **755** | **6653** |

**3.2 ITMA'INN System Overview and General Framework**

Figure 1 presents the overall architecture of the proposed ITMA'INN system for Monkeypox diagnosis and tracking. The workflow can be summarized in the following six key components:

1. **Patient interactions with the mobile application:** Users can access the mobile application as registered users or continue as guests. The patient can upload an image of the suspected skin lesions for Monkeypox detection and optionally input symptoms for analysis.
2. **Data Transmission and Preprocessing:** Uploaded data is transmitted securely to the backend, where image processing techniques are applied on the images to enhance quality and ensure compatibility with the DL-based model.
3. **DL-Based Image Classification:** The preprocessed image is fed into a fine-tuned DL model based on pretrained vision transformers or CNN, which predicts the likelihood of Monkeypox infection (binary or multiclass).
4. **Diagnosis and User Feedback:** The model's prediction, along with the confidence score, is displayed to the user on the application screen. The patients can also receive some recommendations and educational content.
5. **Health Authority Dashboard:** A real-time dashboard, synchronized with the app database, allows public health authorities to monitor Monkeypox case trends, track geographic spread, and support timely interventions.
6. **System Administration:** Backend operations, including performance monitoring, user management, and system maintenance, are handled by the system administrator.

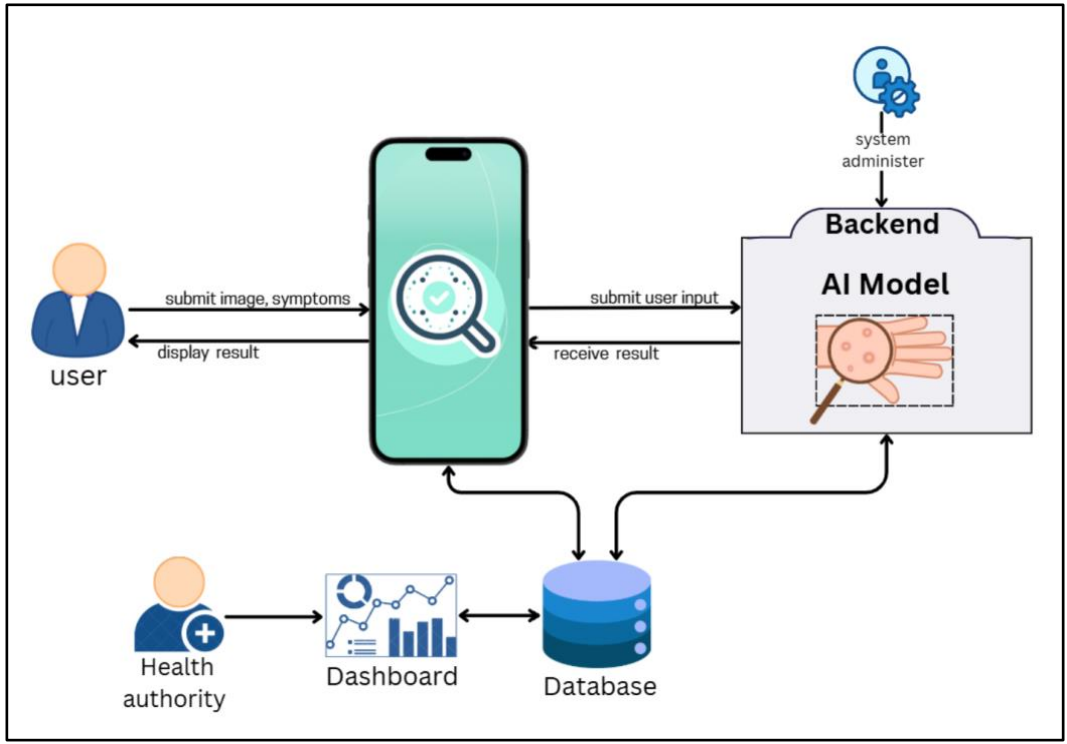

**Fig. 1.** The Workflow of the ITMA'INN system

We explained each of these steps and components in more detail in the corresponding sections.

### 3.3 AI-Model Pipeline

The AI model development pipeline for Monkeypox detection involves four primary stages, each critical to ensuring accurate and reliable classification of skin lesion images. Figure 2 illustrates the pipeline, showing each stage, which can be explained as follows:

The pipeline of building an AI model for Monkeypox detection involves four primary stages, each critical to ensuring accurate and reliable classification of skin lesion images. Figure 2 illustrates the pipeline, showing each stage, which can be explained as follows:

1. **Image Preprocessing and Augmentation:** This initial stage consists of 2 steps. First, preprocessing entails a few crucial steps to enhance image quality and guarantee model compatibility, including image enhancement, resizing (to match model input dimensions), normalization, and optional cropping. Data augmentation is the second step, which is employed to boost the training data's volume and diversity.

2. **Feature Extraction via Pretrained Models and Transfer Learning:** Pretrained DL models are employed to extract relevant features from images automatically. These models, originally trained on large-scale datasets such as ImageNet, offer a significant advantage in terms of saving time and computational resources. Next, these models are adapted to the Monkeypox detection task by fine-tuning selected layers in a transfer learning style.

3. **Classification:** The extracted features are passed to fully connected layers to predict a label to indicate whether the image shows signs of Monkeypox, and the classifier categorizes these predictions into one of the two classes (Monkeypox or Not Monkeypox) in the binary classification. In the multiclass classification scenario, the image is classified into one of six categories (e.g., Monkeypox, Chickenpox, Measles, etc.).

4. **Evaluation and Optimization:** Models are evaluated using validation data and standard metrics. Hyperparameter tuning is employed to enhance model performance and generalization. Detailed experimental settings and metric definitions are explained later in the Experiments and Evaluation section.

5. **Model Deployment:** The best-performing model is selected and deployed within the ITMA'INN mobile application.

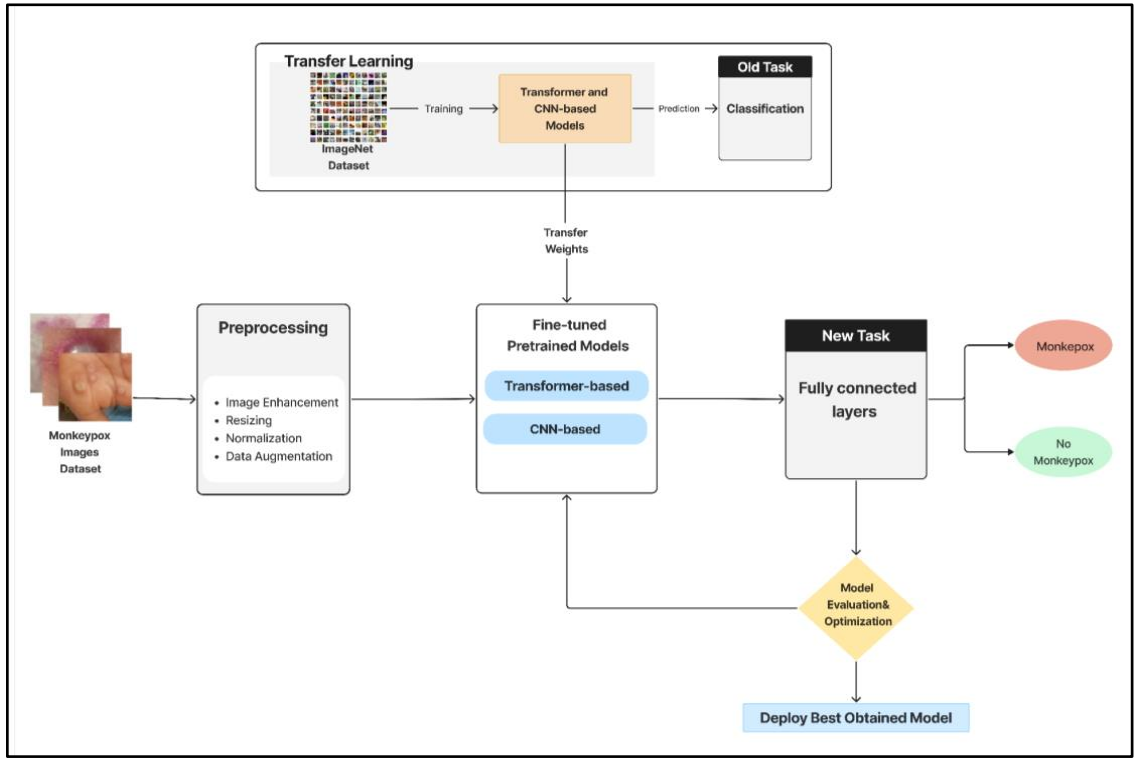

**Fig. 2.** AI-Model Pipeline for Monkeypox Detection in the ITMA'INN System using Binary Classification.

### 3.3.1 Pretrained Models

In this study, we focused primarily on transformer-based and hybrid models, as recent research suggests their superior performance over traditional CNNs in skin lesion classification tasks (Nie et al., 2022; Yolcu Oztel, 2024). Hence, we employed a broad range of models, including ViT, TNT, Swin Transformer, MobileViT, ViT Hybrid, and ResNetViT, to evaluate their performance in Monkeypox detection. Additionally, to establish a strong baseline for comprehensive comparison, we also included the most popular CNN-based models, such as VGG16, ResNet50, and EfficientNet-B0. With this broad model selection, we could assess the relative strengths of modern transformer-based approaches against standard and well-known CNN architectures. A brief overview of each model and its underlying architecture is provided in the subsequent section.

**Vision Transformer (ViT)** (Dosovitskiy et al., 2020; Mao et al., 2021)**.**

ViT is one of the first transformer architectures designed for computer vision tasks. Inspired by the success of transformers in natural language processing (NLP) (Vaswani et al., 2017), several efforts have emerged to apply this success to images. Unlike CNNs, which process images as a whole, ViT divides images into fixed-size patches that serve as tokens similar to the way words are treated in text. These tokens are passed through a standard transformer encoder using self-attention mechanisms to model spatial relationships. ViT relies on data-driven learning, requiring large-scale datasets for training. After the emergence of ViT, several variants have been proposed (Han et al., 2023; Liu et al., 2021; Zhou et al., 2021), with hybrid architectures integrating both characteristics of ViTs and CNNs.

**Transformer in Transformer (TNT)** (Han et al., 2021).

The TNT model extends the standard Vision Transformer by introducing a two-level embedding mechanism that captures local and global representations. Each image is divided into fixed-size patches, which are then further subdivided into a series of pixels. Pixel values are transformed into pixel embeddings via a linear projection to capture the local dependencies. These embeddings are then processed within an inner transformer block. This block's output is combined to create patch embeddings, which are sent to an outer transformer block that records the global interactions between patches. TNT preserves fine-grained local information while learning high-level global features, enhancing performance in several tasks.

**Swin Transformer** (Liu et al., 2021).

Swin Transformer processes images by dividing them into non-overlapping patches, which are passed through hierarchical stages of transformer blocks. A key innovation is the shifted window mechanism, which enables efficient local self-attention while capturing cross-region dependencies. This enables the model to effectively capture relationships between neighboring regions without calculating attention globally. The model constructs a hierarchical architecture, and the number of tokens decreases step by step through patch merging layers, like how pooling is done in CNNs. That leaves the multi-scale feature maps, thereby making the Swin Transformer fit for downstream applications like object detection and segmentation. The model's architecture is scalable in different sizes (e.g., Swin-Tiny, Swin-Base, etc.) and offers linear computational complexity with image size.

**MobileViT** (Mehta and Rastegari, 2021).

MobileViT is a lightweight hybrid architecture that combines the strengths of CNNs and Transformers for mobile-friendly vision tasks. It introduces a novel building block called the MobileViT block, designed to learn both local features (using convolutions) and global representations (using transformers) efficiently. This design enables efficient inference while maintaining high accuracy on resource-constrained devices.

**ViT Hybrid** (Dosovitskiy et al., 2020).

ViT Hybrid extends the standard ViT by using a CNN backbone (e.g., BiT) to extract feature maps, which are then used as input tokens for the transformer encoder. The ViT Hybrid leverages a convolutional backbone, specifically BiT (Big Transfer), to extract feature maps. These convolutional features are used as the initial "tokens" for the transformer encoder. This hybrid approach aims to combine the local spatial inductive bias of CNNs with the global contextual modeling of transformers, resulting in improved learning efficiency and performance on image recognition tasks.

**ResNet - Vision Transformer (ViT) hybrid (ResNetViT)** (Dosovitskiy et al., 2020).

The ResNet-ViT Hybrid model combines a CNN ResNet with a ViT to effectively capture both local and global features from images. The ResNet backbone extracts spatial feature maps, which are then passed to the ViT encoder for global context modeling using self-attention. This hybrid approach leverages the strengths of CNNs (local detail and inductive bias) and transformers (long-range dependency modeling), resulting in a more powerful and generalizable image representation. The model variant we selected does not include a classification head, so we opted to use it as a feature extractor and appended a custom classification head tailored to our Monkeypox detection task.

**VGG16** (Simonyan and Zisserman, 2014).

This deep CNN became popular due to its simplicity and good performance on image classification problems. Its structure has 13 convolutional layers and 3 fully connected layers, totaling 16 weight layers (as the name indicates, VGG16). It uses very small 3×3 convolution filters, stacked in depth to capture complex patterns. These two aspects combined allow the network to learn hierarchical and rich visual features. After convolution and pooling layers, the network ends with two dense layers of 4096 units each, followed by a final classification layer. ReLU activation is used in all hidden layers, and max pooling reduces the spatial dimensions. While straightforward, VGG16 has become the new benchmark for classification accuracy and has since become a widely used feature extractor in transfer learning across many vision tasks.

**ResNet50** (He et al., 2015)**.**

ResNet, or Residual Network, is a deep CNN structure that led to the development of residual learning to make it possible to train very deep networks to be practical and effective. Its key innovation lies in the residual block, which introduced a shortcut (or skip) connection that bypasses one or more layers. Instead of learning the entire output directly, each block learns a residual function (i.e., i.e., the difference between the input and the output) that make the optimization process easier especially for very deep architectures. This design solves the vanishing gradient issue and the degradation problem (where the deeper models perform less well than the shallower models because of optimization issues). As a result, ResNet models, including ResNet50 (with 50 layers), can be scaled to hundreds, and thousands of layers, and still successfully train.

**EfficientNet-B0** (Tan and Le, 2019).

EfficientNet is a family of convolutional neural networks designed to achieve high accuracy with optimal computational efficiency. It was created with neural architecture search and built predominantly using mobile inverted bottleneck Convolutions (MBConv) blocks, similar to those in MobileNetV2, but enhanced with Squeeze-and-Excitation (SE) blocks for better feature recalibration. A key innovation in EfficientNet is compound scaling, where the model depth, width, and input resolution are all uniformly scaled by constant coefficients. Such a balanced treatment yields significantly better performance than previous models scaling a single dimension.

EfficientNet-B0, the baseline model, is lightweight and efficient, with only 5.3 million parameters and 0.39 billion FLOPs, yet it achieves 77.1% top-1 accuracy on ImageNet, outperforming much larger models like ResNet50.

**3.3.2 Fine-tuning and Transfer Learning**

To adapt pretrained models to the Monkeypox image classification task, we applied transfer learning, a popular DL approach that leverages pretrained models to solve new, related problems with limited data. All selected models were initially pretrained on large-scale datasets such as ImageNet-1k or ImageNet-21k, which enabled them to learn strong, general-purpose visual features. We fine-tuned these models by replacing their original classification heads with custom fully connected layers suited to our binary or multiclass classification objectives. To reduce the risk of overfitting and improve training efficiency, we typically froze most of the pretrained layers, particularly the feature extraction blocks, and trained only the new classification head and a few top layers. For transformer-based models such as MobileViT and ViT-Hybrid, we froze the early transformer blocks and fine-tuned only the final transformer layers, including the classification head.

All models were implemented using the PyTorch framework (v2.0.1) and the Python programming language (v3.10.12). Additional libraries were utilized, including Torchvision for data handling and transformations, Timm (PyTorch Image Models) for loading various pretrained transformer architectures, scikit-learn for performance metrics such as precision, recall, and F1-score, Torchinfo for model summarization, and Matplotlib for plotting training curves. The experiments were conducted on the Google Colab Pro using NVIDIA Tesla T4 GPUs and extended high-RAM environments to support successful training and testing. Key steps in the Code-level configurations and pipeline included:

- Loading the pretrained weights for each model using timm.create_model.
- Freezing most of the backbone layers to preserve general feature representations and fine-tuning the final layers to learn from the Monkeypox dataset.
- Replacing the classification head with custom fully connected layers for binary or multiclass classification tasks, followed by a Sigmoid activation function for binary classification and a SoftMax activation function for multiclass classification.
- Setting random seeds for reproducibility across multiple experiments.
- Applying classical data augmentation techniques such as resizing, normalization, and interpolation via timm.data.create_transform to improve generalization.
- Using early stopping and other regularization techniques to prevent overfitting based on validation loss.
- Hyperparameter optimization was conducted by tuning several key parameters, with selected tested values summarized in Table 3. Key hyperparameters such as the learning rate, batch size, optimizer, dropout rate, and weight decay were adjusted repeatedly through multiple training iterations.

| Hyperparameters | Values |
|---|---|
| Learning rate | 0.001, 0.0001, **1e-5**, 2e-5 |
| Batch size | 8, **16**, 32 |
| Dropout rate | 0.2, **0.3**, 0.4 |
| Weight decay | 1e-4, 1e-5 |
| Optimizer | **Adam**, AdamW |

**Table 3** Summary of hyperparameters optimized during the training. Bold indicates the selected values.

## 5. Experiments and Evaluation Protocols

The experimental setup for evaluating the AI models implemented in the ITMA'INN system focused on assessing the performance of multiple pretrained DL models in detecting and classifying Monkeypox skin lesions from images using several approaches and multiple evaluation metrics.

### 5.1 Experiments Setting

We conducted two separate experiments: one for binary classification (Monkeypox vs. non-monkeypox) and another for multiclass classification (Monkeypox, Chickenpox, Measles, Cowpox, HFMD, and Healthy skin). Both experiments followed the same evaluation protocols and utilized the same evaluation metrics.

For each experiment, we applied two data splitting approaches:

1. **Random 80:20 train-test split in a stratified style**, where each subset includes the same percentage of each class label.
2. **Stratified 5-fold cross-validation (CV)**, where the data was randomly split into five subsets. In each iteration, one subset was reserved for testing, while the remaining four were used for training. This procedure was repeated five times, ensuring each subset served once as the test set. The results were averaged across all folds to obtain a more reliable and generalizable estimation of model performance.

Training and testing processes were implemented systematically with custom functions to track key metrics at every epoch. Such modular implementation allowed to systematically fine-tune and test a broad spectrum of models and thus ensured consistency and reproducibility across all models and experiments.

### 5.2 Evaluation Metrics

To evaluate the prediction performance of the AI models, several evaluation metrics were calculated. These metrics provide a comprehensive view of the model's classification performance suited for both binary and multiclass problems. Table 4 summarizes the definitions and corresponding mathematical formulas for each metric. Most of these metrics (Davis and Goadrich, 2006; Martinez-Ramon et al., 2024) are defined mathematically and is derived from the confusion matrix components: true positives (TP), false positives (FP), true negatives (TN), false negatives (FN). For all performance metrics, higher values (closer to one) indicate better performance, except for the loss metric, where lower values represent better model performance.

**Table 4** Performance Evaluation Metrics of the DL pretrained models.

| Metrics | Definition | Mathematical Formula |
|---|---|---|
| Accuracy (Acc) | It is the ratio of correctly classified samples to the total number of samples in the dataset. | $Acc = (TP + TN)/(TP + TN + FP + FN)$ |
| Precision (PR) | The proportion of true positive predictions among all positive predictions, reflecting the model's ability to avoid false positives. | $PR = TP/(TP + FP)$ |
| Recall (RC) | The proportion of true positives among all actual positives, indicating how well the model captures relevant cases. | $RC = TP/(TP + FN)$ |
| F1-score | The harmonic mean of precision and recall, providing a balance between the two. | $F1\text{-}score = (2 \cdot PR \cdot RC)/(PR + RC)$ |
| AUC-ROC | Is the area under the ROC curve, measuring the model's ability to distinguish between classes | $AUC = \sum_{i=1}^{n-1}(FPR_{i+1} - FPR_i) \cdot \frac{TPR_{i+1}+TPR_i}{2}$ |
| Loss | It measures how well the model's predictions match the actual outcomes, representing the model's prediction error. | $MSE = \frac{1}{N}\sum_{i=1}^{N}(y_i - \hat{y}_i)$ |

## 6. Results and Discussion

In this section, we present and analyze the experimental results obtained using both binary and multiclass classification datasets under two data splitting strategies: 80/20 train-test split and 5-fold cross-validation. We evaluate and compare the performance of several transformer-based and CNN-based pretrained models. We further compared our model's performance with the best results of selected state-of-the-art methods and discussed our model's key strengths that might enhance our model's performance.

**6.1 AI Model Prediction Performance**

We evaluated the performance of multiple pretrained models on Binary and multiclass classification tasks.

**6.1.1 Binary Classification Task**

As shown in Table 5, the highest test accuracy of 97.78% was achieved by VGG16, ViT, TNT, and MobileViT, all demonstrating outstanding overall performance. Three transform-based models (ViT, MobileViT, and TNT) managed to attain high precision and F1-score, supported by robust AUC scores (0.9800), which reflect their high discriminative capabilities. ResNetViT, Swin Transformer, and ViT Hybrid models obtained competitive accuracies of 95.56% and AUCs of 0.9600, reflecting their satisfactory but slightly lower performance compared to the transformer-based models alone. On the other hand, models like ResNet50 and EfficientNet-B0 attained lower accuracies of 91.11% and 93.33%, respectively, albeit with marginally higher loss values and lower AUCs, indicating that traditional CNN-based models, as good as they were, were trumped by transformer-based or hybrid models in this binary classification task.

Additionally, we applied 5-fold CV using the three best-performing pretrained models to validate their robustness further. While the results were lower compared to the 80:20 train-test split, the models still demonstrated reasonable performance, achieving approximately 89% accuracy and F1-score. This decline in performance can be attributed to the reduced training and testing set sizes per fold, which emphasizes a key limitation of using transformer-based pretrained models on small datasets. These findings suggest that such models benefit significantly from larger datasets and that applying CV settings effectively may require a more extensive collection of labeled Monkeypox images. Overall, the prediction performance shows that transformer and hybrid transformer models, especially ViT, TNT, and MobileViT, are extremely efficient at Monkeypox image classification, beating conventional convolutional architectures in most of the evaluation metrics.

**Table 5:** Prediction Performance of Several Pretrained Models on the Binary Classification Task.
**Bold** font indicates the best-performing models and *Italic* font indicates the second-best performing models.

| Model name | Accuracy | Precision | F1-score | Recall | Loss | AUC |
|---|---|---|---|---|---|---|
| 80:20 Train-Test Split ||||||| 
| *ResNetViT* | *0.9556* | *0.9596* | *0.9557* | *0.9556* | *0.2355* | *0.9600* |
| ResNet50 | 0.9111 | 0.9152 | 0.9114 | 0.9111 | 0.2575 | 0.9150 |
| **TNT** | **0.9778** | **0.9788** | **0.9778** | **0.9778** | **0.1729** | **0.9800** |
| *Swin Transformer* | *0.9556* | *0.9596* | *0.9557* | *0.9556* | *0.1770* | *0.9600* |
| **ViT** | **0.9778** | **0.9788** | **0.9778** | **0.9778** | **0.1669** | **0.9800** |
| **VGG16** | **0.9778** | **0.9788** | **0.9778** | **0.9778** | **0.1030** | **0.9800** |
| *ViT Hybrid* | *0.9556* | *0.9596* | *0.9557* | *0.9556* | *0.1656* | *0.9600* |
| **MobileViT** | **0.9778** | **0.9786** | **0.9777** | **0.9788** | **0.1967** | **0.9750** |
| EfficientNet-B0 | 0.9333 | 0.9345 | 0.9335 | 0.9333 | 0.2969 | 0.9720 |
| 5-Fold Cross-validation |||||||
| ViT | 0.8903 | 0.8914 | 0.8898 | 0.8903 | 0.4701 | 0.8871 |
| MobileViT | 0.8773 | 0.8783 | 0.8768 | 0.8773 | 0.4133 | 0.8731 |
| TNT | 0.8990 | 0.9022 | 0.8984 | 0.8990 | 0.5395 | 0.8944 |

**6.1.2 Multiclass Classification Task**

The same was applied to multiclass classification. Table 6 shows that the ViT Hybrid and ResNetViT models achieved the highest test accuracies of 92.16%, with very high precision, recall, and F1-scores reflecting their superiority in handling multiclass predictions. ViTHybrid performed best with the lowest loss value of 0.2501 among all the models, while ResNetViT achieved a very high AUC of 0.9823, reflecting good discrimination between multiple classes. EfficientNet-B0 also performed well, with a test accuracy of 89.54% and an extremely high AUC of 0.9797, demonstrating its balance between efficiency and accuracy even in more difficult classification tasks. MobileViT and Swin Transformer followed with accuracies above 84%. ViT, TNT, VGG16, and ResNet50 lagged behind, with accuracies between 80.39% and 84.31%, and higher loss values. We can also observe that ViT, TNT, VGG16, and ResNet50 performed comparatively worse, with accuracies ranging from 80.39% to 84.31% and greater loss values, particularly that of VGG16 (0.7762) and ResNet50 (0.6292).

These results imply that although transformer-based models are commonly strong candidates for multiclass classification, their performance is influenced by architectural variations and fine-tuning strategies. The findings highlight that hybrid and transformer-based architectures, mainly ViT Hybrid and ResNetViT, consistently outperform conventional CNN-based models in multiclass skin lesions classification tasks.

**Table 6**: Prediction Performance of Several Pretrained Models on the Multiclass Skin Lesion Classification Task. **Bold** font indicates the best-performing models and *Italic* font indicates the second-best performing model.

| Model name | Accuracy | Precision | F1-score | Recall | Loss | AUC |
|---|---|---|---|---|---|---|
| 80:20 Train-Test Split | | | | | | |
| **ResNetViT** | **0.9216** | **0.9271** | **0.9224** | **0.9216** | **0.3079** | **0.9823** |
| ResNet50 | 0.8039 | 0.8169 | 0.8043 | 0.8039 | 0.6292 | 0.9509 |
| TNT | 0.8170 | 0.8301 | 0.8157 | 0.8170 | 0.4797 | 0.9611 |
| Swin Transformer | 0.8497 | 0.8596 | 0.8503 | 0.8497 | 0.4288 | 0.9736 |
| ViT | 0.8431 | 0.8551 | 0.8413 | 0.8431 | 0.5175 | 0.9655 |
| VGG16 | 0.8039 | 0.8062 | 0.8046 | 0.8039 | 0.7762 | 0.9570 |
| **ViT Hybrid** | **0.9216** | **0.9262** | **0.9219** | **0.9216** | **0.2501** | **0.9286** |
| MobileViT | 0.8758 | 0.8782 | 0.8746 | 0.8758 | 0.4905 | 0.9730 |
| *EfficientNet-B0* | *0.8954* | *0.9002* | *0.8961* | *0.8954* | *0.3559* | *0.9797* |
| 5-Fold Cross-validation | | | | | | |
| ResNetViT | 0.85288 | 0.85734 | 0.85188 | 0.85288 | 0.45666 | 0.9665 |
| ViT Hybrid | 0.8388 | 0.84282 | 0.83604 | 0.8388 | 0.45642 | 0.96602 |

## 6.2 Comparison with the State-of-the-art Methods

To evaluate the effectiveness of our proposed system, we benchmarked the best-performing models against prior state-of-the-art methods on both binary and multiclass Monkeypox classification tasks using the same public dataset versions (MSLD and MSLD v2.0).

Table 7 presents a comparison with two studies focused exclusively on binary classification tasks (Monkeypox vs. Non-Monkeypox) using the same version of the dataset we utilized in our study (MSLD). These studies include Sahin et al. (Sahin et al., 2022) and Nayak et al. (Nayak et al., 2023), both of which used traditional CNN architectures such as MobileNetV2, GoogLeNet, and ResNet-18. Our MobileViT model outperformed the MobileNetV2-based approach presented by ((Sahin et al., 2022), which reported an accuracy of 91.11%. Our MobileViT achieved a remarkably higher accuracy of 97.78%. This represents a 6.67 percentage point improvement in classification accuracy, demonstrating the added value of transformer-based hybrid architectures over traditional CNNs for medical diagnosis tasks like Monkeypox detection. This result underscores the effectiveness of MobileViT's hybrid architecture, which integrates local feature learning from convolutions with global context modeling through transformers, a key advancement over the traditional MobileNet.

We also compared our model with the work by Nayak et al. (Nayak et al., 2023) who reported several pretrained model results but we picked the best two (with the best round results) for direct comparison. GoogLeNet delivered competitive results across most evaluation metrics; however, its recall (sensitivity) was notably lower at 94.12%, indicating that 5.88% of actual Monkeypox cases were missed. In healthcare context, missing positive cases can be risky, as it may lead to false reassurance, delayed treatment, and increased disease transmission. On the other hand, another pretrained model proposed by the same study (Nayak et al., 2023), ResNet-18, achieved slightly higher results in four evaluation metrics. However, the study lacks details on data splitting protocols, which raises concerns about overfitting. Moreover, it remains a relatively heavier model in terms of structure and computational complexity. In contrast, our use of MobileViT, designed to be lightweight and mobile-friendly—achieved a strong accuracy of 97.78%, with balanced precision and recall. Its hybrid architecture, combining convolutional operations with transformer mechanisms, enabled the effective extraction of both local and global features. Trained on an augmented dataset of 2,607 images, MobileViT demonstrated robust generalization and MobileViT and provided an optimal trade-off between accuracy and model efficiency, making it well-suited for deployment in real-time mobile health systems. Its hybrid design, combining convolutional and transformer layers, enabled effective feature extraction of both local and global patterns from skin lesion images, which is crucial for visual diagnosis.

**Table 7** Comparison with Binary Classification Studies Using the MSLD Dataset. **Bold** indicates the best-performing models and *Italic* font indicates the second-best performing model.

| Study | Model | Accuracy (%) | Precision (%) | Recall (%) | F1-Score (%) | AUC (%) |
|---|---|---|---|---|---|---|
| (Sahin et al., 2022) | MobileNetv2 | 91.11% | 90.00% | 90.00% | 90.00% | - |
| (Nayak et al., 2023) | ResNet-18 | **99.49%** | **98.52%** | **99.44%** | **99.49%** | - |
| | GoogLeNet | 97.37% | 97.46% | 94.12% | 97.05% | - |
| Proposed system: **ITMA'INN** | MobileViT | *97.77%* | *97.86%* | *97.88%* | *97.77%* | *97.50%* |

For multiclass classification, Table 8 presents a comparison with the only recent study that applied classification on the same skin lesion dataset (MSLD v2.0), which includes six classes: Monkeypox, Chickenpox, Measles, Cowpox, HFMD, and Healthy skin (Ali et al., 2024). While both studies utilized the same dataset, the methodologies differ notably. Ali and coauthors evaluated several CNN-based models and applied standard and color-space augmentation techniques, achieving their best performance with DenseNet121 at an accuracy of 83.59%. In contrast, our approach applied a broader set of augmentation techniques and adopted transformer-based architectures, leading to better classification results and stronger generalization by outperforming DenseNet121 by 8.57% in terms of accuracy. Moreover, the successful classification across six visually similar skin conditions demonstrates the robustness of our system, addressing a diagnostic challenge even for trained clinicians.

**Table 8** Comparison with Multiclass Classification Studies Using the MSLDv2.0 Dataset. **Bold** indicates the best-performing models

| Study | Model | Accuracy (%) | Precision (%) | Recall (%) | F1-Score (%) | AUC (%) |
|---|---|---|---|---|---|---|
| (Ali et al., 2024) | DenseNet121 | 83.59% | 85% | - | - | - |
| *Proposed system: ITMA'INN* | **ResNetViT** | **92.16%** | **92.71%** | **92.16%** | **92.24%** | **98.23%** |

Overall, our approach demonstrates the advantages of transfer learning, pretrained model fine-tuning, and careful hyperparameter optimization. By incorporating these strategies, we successfully adapted powerful pretrained models to the domain of skin lesion classification, achieving state-of-the-art outcomes while preserving the computational efficiency needed for clinical and mobile applications.

## 6. Mobile Application and Dashboard Development and Integration

After finalizing and evaluating the AI model pipeline for Monkeypox detection and to demonstrate the practical use of our model, we moved to the next phase which is integrating the best-performing, mobile-compatible model into a functional system accessible to end users (i.e., patients). This was done by developing a cross-platform mobile application and an administrative monitoring dashboard.

### 6.1 Mobile Application Design and Testing

We created the ITMA'INN mobile application to provide an intelligent and user-friendly diagnostic tool for the identification of Monkeypox. The application also allows patients to interact with the system, access relevant features, and receive appropriate guidance and support. The Flutter framework was used to guarantee cross-platform compatibility on both iOS and Android. The app's backend is hosted on Firebase, leveraging services such as Cloud Functions, Firestore for real-time database operations, and Firebase Authentication for secure user management.

After development, we tested the ITMA'INN app to ensure reliable performance and a user-friendly experience. When users launch the ITMA'INN application for the first time, they are guided through a brief onboarding process before accessing the main diagnostic functionality. The core interaction begins at the Examine Screen, where users submit symptom descriptions and upload skin lesion images from their camera, photo library, or the device files, for analysis. Based on the AI model's prediction, users are directed either to the Infected Screen or the Uninfected Screen. If the user is determined to be infected, the system provides a list of nearby health centers using location services to facilitate prompt medical follow-up. This workflow was tested to ensure reliability in user navigation, accurate result display, and effective health center recommendations based on geolocation data. Figure 3 displays the main user interfaces of the ITMA'INN application, divided into two halves. The top row presents the interaction flow for users predicted to have Monkeypox. It begins on the Examine Screen, where the user uploads a photo of their skin lesion and selects symptoms. After entering the details, they are taken to the Results Screen, where the AI model prediction and a "Show Nearest Health Center" button are displayed. On clicking this, a screen with a list of health centers near the user's geographic location is displayed.

The bottom row shows the corresponding interaction flow for users who are not predicted to be infected. Such users go through the same first steps, image upload and symptoms selection, but are ultimately taken to a screen saying that no signs of Monkeypox were detected.

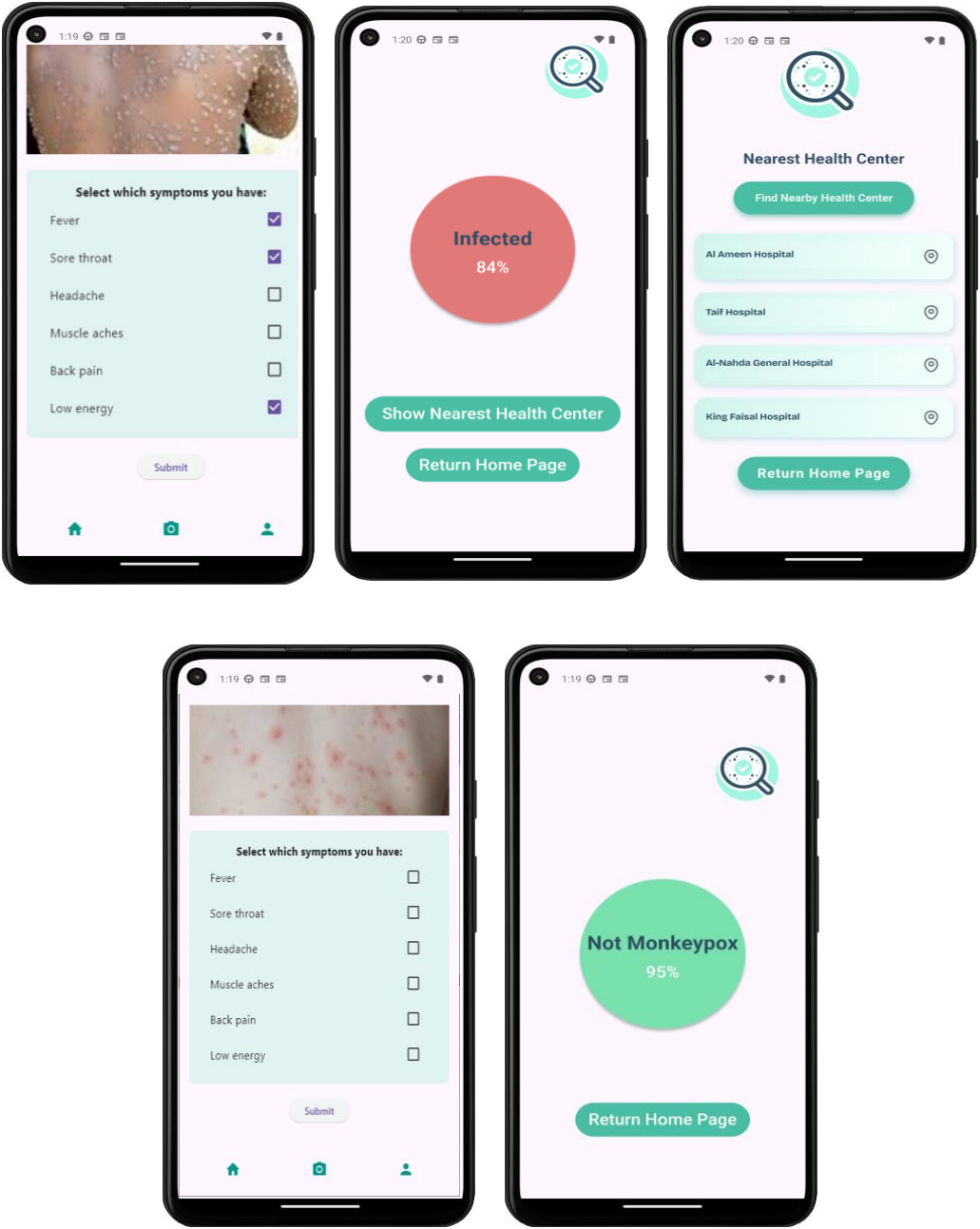

**Fig. 3.** ITMA'INN Application Testing Scenario

**6.2 Monitoring Dashboard Design and Testing**

As part of the system implementation process, a real-time interactive dashboard was developed for the health authorities to monitor Monkeypox cases. The dashboard displays visualization charts based on data collected from the patients of the ITMA'INN mobile application. It shows key metrics, including the number of infected and non-infected Monkeypox patients, infection rate, patient age and gender, symptom presence, and geographical distribution of infected cases. As shown in Figure 4, the dashboard data flow begins with the ITMA'INN mobile application, where patient data is stored in the Firebase database. This data is then streamed to Google BigQuery, where the raw data is organized and analyzed. Power BI is then utilized as a visualization tool, linked to Google BigQuery through a DirectQuery connection to ensure real-time data updates. To make the dashboard accessible, a Power BI Service report is embedded in the ITMA'INN web page portal. The portal uses Power BI authentication, which prompts the health authority user to sign in with their organization's credentials before accessing the dashboard, which ensures secure access.

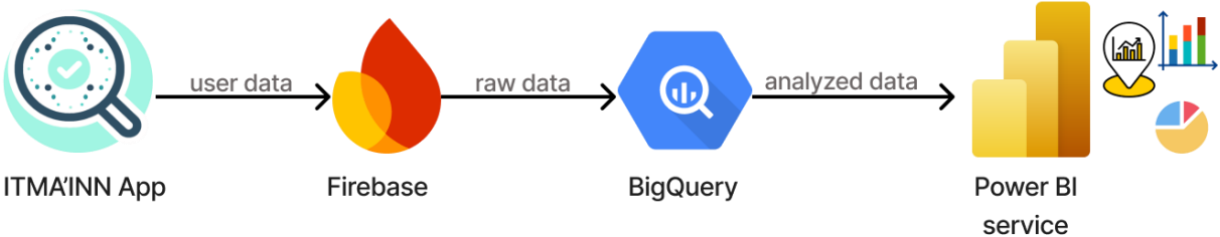

**Fig. 4.** The data flow supporting the Monkeypox Monitoring Dashboard

The **dashboard** (see Figure 5) provides administrators with a secure, real-time Monkeypox monitoring system, starting with the **Dashboard Login Screen**, which requires the organization's email authentication to access the analytics portal. Once logged in, the **Dashboard Overview Screen** displays critical metrics like infection rates (e.g., 57%), demographic breakdowns (e.g., 75% male cases), and a detailed patient table with symptoms, test dates, and locations, all updated in real-time. Administrators can track outbreaks, analyze trends, and make data-driven decisions, while the system ensures security through strict credential checks and session management. Testing focuses on login validation, data accuracy, real-time updates, and responsiveness, ensuring reliable performance for public health monitoring.

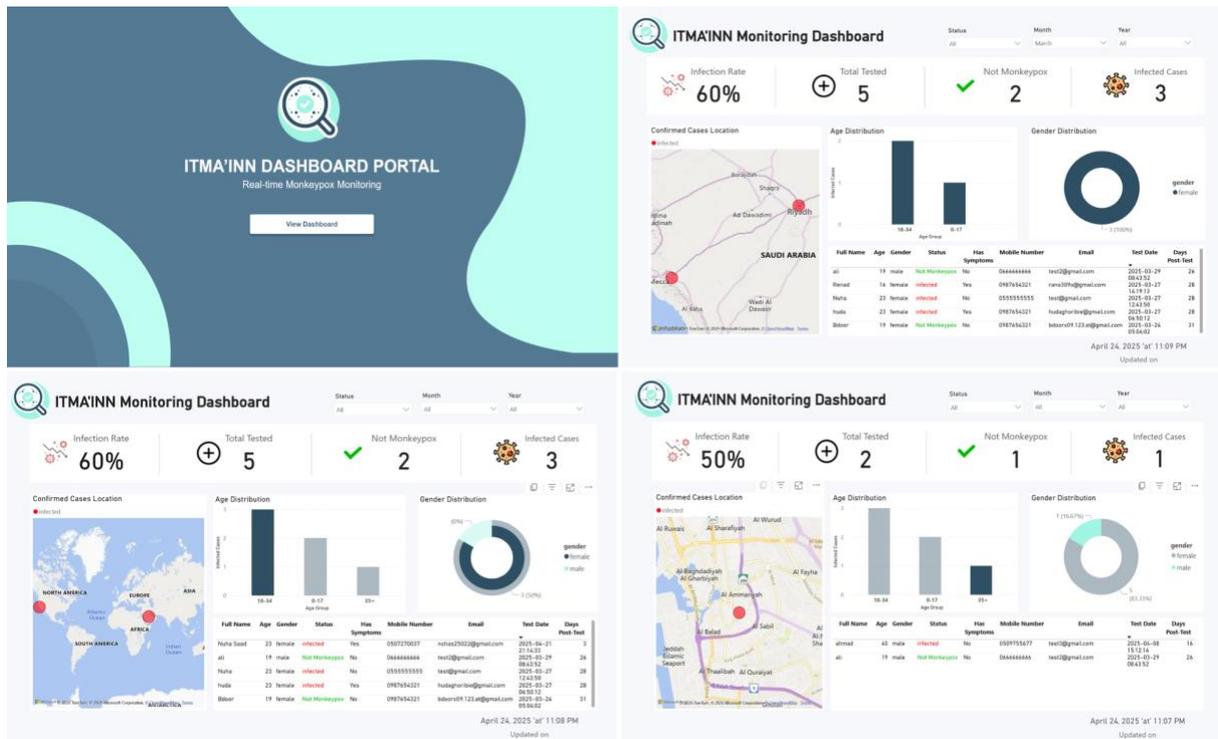

**Fig. 5.** Overview of the monitoring dashboard in the ITMA'INN system.

## 7. Conclusion

This paper proposed the design and development of *ITMA'INN*, an intelligent, AI-powered mobile healthcare system for the early detection and monitoring of Monkeypox from skin lesion images. By integrating deep learning pretrained models into a user-friendly mobile application, the system enables the patient to capture and upload skin lesion images, respond to symptom-related questions, send safety messages, receive diagnostic results, and access nearby health centers, all in real-time. Moreover, a monitoring dashboard was developed to support health authorities.

      The experimental results demonstrated the effectiveness of transformer-based and hybrid-based pretrained models, with MobileViT and ViTHybrid achieving high accuracy in both binary and multiclass classification tasks. The outstanding prediction performance emphasizes its potential for use in mobile health environments, especially to support smart city healthcare initiatives and public health surveillance. Through this work, we have demonstrated the potential for smart health solutions to aid in monitoring and controlling infectious diseases, marking a significant step toward more proactive healthcare management.

      For future enhancement, we aim to expand the scope of diagnosis to include other skin diseases using larger and more diverse datasets, which will contribute to early detection and improved diagnostic accuracy. We also plan to enhance multiclass classification performance through additional model training and fine-tuning to support broader diagnostic capabilities. By implementing these future advancements, ITMA'INN system will be enhanced, contributing not only to the early detection of Monkeypox but also to the broader prevention of other infectious diseases.

## Data Availability

The datasets used for both binary and multiclass classification tasks are publicly available through the Kaggle repository. The links are as follow:
- MSLD Dataset: https://www.kaggle.com/datasets/nafin59/monkeypox-skin-lesion-dataset
- MSLDv2.0 Dataset: https://www.kaggle.com/datasets/joydippaul/mpox-skin-lesion-dataset-version-20-msld-v20

## Competing interests

The authors have declared that no conflict of interests exist.